\documentclass[sigconf]{acmart}
\usepackage{booktabs} % For formal tables
\newcommand{\vect}[1]{\mathbf{#1}}
\usepackage{amsmath}
\usepackage{xcolor}
\usepackage{natbib}

\newcounter{enum}
\newenvironment{packed_enum}{
	\begin{list}{\textbf{(\arabic{enum})}}{
			\setlength{\itemsep}{0pt}
			\setlength{\parskip}{0pt}
			\setlength{\labelwidth}{-5 pt}
			\setlength{\leftmargin}{0 pt}
			\setlength{\itemindent}{0pt}
			\usecounter{enum}}
	}{\end{list}}

\renewcommand\footnotetextcopyrightpermission[1]{} % removes footnote with conference information in first column
\begin{document}
\title{AI-CARGO: A Data-Driven Air-Cargo Revenue Management System}

\author{Stefano Giovanni Rizzo, Ji Lucas, Zoi Kaoudi, Jorge-Arnulfo Quiane-Ruiz, Sanjay Chawla} 
%\authornote{Dr.~Trovato insisted his name be first.}
\affiliation{%
  \institution{Qatar Computing Research Institute (QCRI)}
  %\streetaddress{P.O. Box 1212}
  \city{Doha}
  \state{Qatar}
}
\email{{strizzo,jlucas,zkaoudi, jquianeruiz,schawla}@hbku.edu.qa}
\iffalse
\author{G.K.M. Tobin}
\authornote{The secretary disavows any knowledge of this author's actions.}
\affiliation{%
  \institution{Institute for Clarity in Documentation}
  \streetaddress{P.O. Box 1212}
  \city{Dublin}
  \state{Ohio}
  \postcode{43017-6221}
}
\email{webmaster@marysville-ohio.com}

\author{Lars Th{\o}rv{\"a}ld}
\authornote{This author is the
  one who did all the really hard work.}
\affiliation{%
  \institution{The Th{\o}rv{\"a}ld Group}
  \streetaddress{1 Th{\o}rv{\"a}ld Circle}
  \city{Hekla}
  \country{Iceland}}
\email{larst@affiliation.org}

\author{Valerie B\'eranger}
\affiliation{%
  \institution{Inria Paris-Rocquencourt}
  \city{Rocquencourt}
  \country{France}
}
\fi

\renewcommand{\shortauthors}{B. Trovato et al.}

\begin{abstract}
We propose AI-CARGO, a revenue management system for air-cargo that
combines machine learning prediction with decision-making using mathematical optimization
methods.  AI-CARGO
addresses a problem that is unique to the air-cargo business, namely
the wide discrepancy between the quantity (weight or volume) that a shipper
will book and the actual received amount at departure time by the airline.
The discrepancy results in  sub-optimal and inefficient behavior by
both the shipper and the airline resulting in overall loss of potential 
revenue for the airline. AI-CARGO also includes a data cleaning
component to deal with the heterogeneous forms in which booking
data is transmitted to the airline cargo system.
AI-CARGO is deployed in the production environment of a large commercial
airline company.
We have validated the benefits of AI-CARGO using real and synthetic datasets.
Especially, we have carried out simulations using dynamic programming
techniques to elicit the impact on offloading costs and revenue 
generation of our proposed system. Our results suggest that combining prediction within a decision-making framework
can help dramatically to reduce offloading costs and optimize revenue generation.

\end{abstract}

\iffalse
\begin{CCSXML}
<ccs2012>
 <concept>
  <concept_id>10010520.10010553.10010562</concept_id>
  <concept_desc>Computer systems organization~Embedded systems</concept_desc>
  <concept_significance>500</concept_significance>
 </concept>
 <concept>
  <concept_id>10010520.10010575.10010755</concept_id>
  <concept_desc>Computer systems organization~Redundancy</concept_desc>
  <concept_significance>300</concept_significance>
 </concept>
 <concept>
  <concept_id>10010520.10010553.10010554</concept_id>
  <concept_desc>Computer systems organization~Robotics</concept_desc>
  <concept_significance>100</concept_significance>
 </concept>
 <concept>
  <concept_id>10003033.10003083.10003095</concept_id>
  <concept_desc>Networks~Network reliability</concept_desc>
  <concept_significance>100</concept_significance>
 </concept>
</ccs2012>
\end{CCSXML}

\ccsdesc[500]{Computer systems organization~Embedded systems}
\ccsdesc[300]{Computer systems organization~Redundancy}
\ccsdesc{Computer systems organization~Robotics}
\ccsdesc[100]{Networks~Network reliability}

\keywords{ACM proceedings, \LaTeX, text tagging}
\fi

\maketitle

%!TEX root = main.tex

\section{Introduction}
\begin{figure}[t!]
\includegraphics[width=0.3\textwidth]{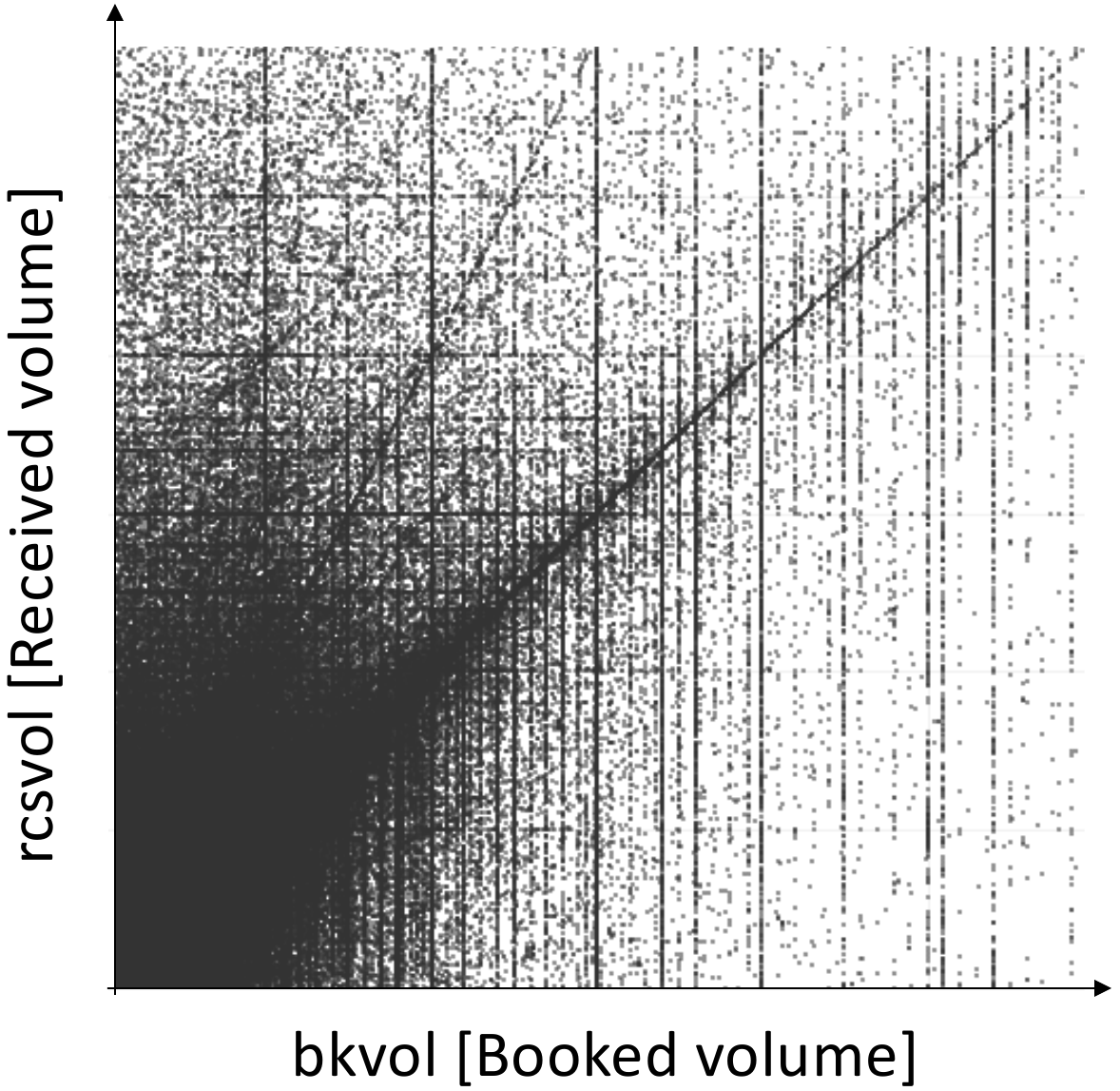}
\caption{The x-axis is booked volume (bkvol).
The y-axis is received vol (rcsvol).
%The AI-CARGO system predicts the rcsvol from bkvol and other features and uses the prediction accept or reject a booking in order to maximize revenue and reduce offloading cost.
%The vertical streaks are potential Disguised Missing Values (DMVs)
%that have to be identified and removed to create a more accurate prediction function.
For confidentiality reason, the axis range is obfuscated.}
\label{brvol}
\vspace{-0.4cm}
\end{figure}

The revenue of commercial airlines is primarily derived from sales of passenger tickets and cargo (freight) shipments.
While most modern airlines have implemented sophisticated data-driven passenger revenue management systems, for cargo the situation is different.
The air-cargo ecosystem is complex and involves several players including shippers, freight forwarders, airline- and end-customers.
%The price points are not clear, their is asymmetric information and the market for air cargo is not efficient.
Overall, there are five fundamental differences between passenger and cargo revenue management~\cite{popescu2006air}\cite{boonekamp2013air}\cite{amaruchkul2007single}:

\begin{packed_enum}
\item In the case of passenger revenue, the unit of sale is an airline seat, which is static.
However, in the case of cargo, there is substantial variability in both volume and weight of cargo shipments.
Furthermore, the revenue from a cargo shipment often depends on the nature of the cargo.
For example, perishable and non-perishable shipments generate different marginal revenues.
This makes the unit of sale in cargo highly dynamic.

\item A large chunk of air cargo capacity is pre-booked by freight forwarders who tend to overbook and release capacity closer to the date of departure.
The ecosystem of air-cargo management is such that there is no penalty for overbooking.
Additionally, some portion of cargo space is also reserved for mail and passenger baggage.
Thus, the effective capacity available for cargo is called the ``free sale'', which can vary up to departure day.

\item For cargo shipments what matters is the source and destination.
How the cargo is routed from source to destination is less of a concern as long as it reaches on time.
Rerouting though has ancillary cost as the shipment has to be stored in a warehouse.

\item A unique aspect of the air-cargo ecosystem is that there is often a substantial discrepancy between the space booked by a shipping agent (in terms of volume and weight) for a particular item and the actual quantity that arrives on or just before the departure day. 
Furthermore, it is  a convention in the  business that airlines will not charge for the discrepancy, which makes it very difficult for airlines to manage their perishable capacity.
This creates an inefficiency in the market where agents tend to book excess capacity.
Therefore, airlines tend to overbook flights under the assumption that the quantity that will arrive will be less than what was booked.
Overbooking often leads to offloading, which has cost in terms of storage and rerouting.
%Even in the case of free-sale there is a substantial difference in the amount booked and what is actually received.

\item 
It is well known that cargo capacity is often volume-constrained, i.e.,~the aircraft will reach volume capacity before it reaches weight capacity. However, this makes things even harder because the volume measurements are less accurate than the weight ones.
Figure~\ref{brvol} illustrates this fact.
\end{packed_enum}

Due to these major differences, air-cargo business requires not only to accurately predict the quantity (weight and volume) of an item that will be tendered but also to make decisions on whether to accept or reject a booking for a certain flight. This will enable the airline to greatly increase the efficiency of capacity utilization.
However, prediction and decision-making in an air-cargo setting is non-trivial because of three main reasons.
First, the quality of cargo booking data varies a lot.
Shippers often send information using text messages, emails, spreadsheets, voice calls, or even intermediate their bookings through freight forwarders.
%There is no web information management system where shippers can directly upload their data, 
Second, there are no good features for predicting the final quantity that will be received.
Employees working in the cargo revenue teams use their intuition to decide whether to accept a shipment for a flight or to reroute through another flight.
Third, offloading and rerouting cargo incur high costs in terms of storage and rerouting, which somehow constraints the decision-making.

We present the AI-CARGO system that deals with the above challenges.
Although the use of formal decision-making is well established in the revenue management community~\cite{chiang}, AI-CARGO is the first work that puts all pieces together to provide a complete pipeline for the air-cargo revenue management problem: given an incoming booking it
(i)~identifies if there might exist a substantial difference between the booked volume and the one that might be tendered,
(ii)~predicts the volume that will be tendered, and
(iii)~considers such a volume prediction to make an acceptance/rejection suggestion.
AI-CARGO is currently deployed in production in a large airline company\footnote{We omit the company name for confidentiality reasons.}.
In particular, after surveying related work in Section~\ref{section:relwork}, we make the following contributions in Section~\ref{section:system}:
%For example markov decision
%processes (MDPs) are used in cargo management to decide whether a shipment is to be accepted for a given flight or
%rerouted through another flight. Upto know MDPs use the average volume of a shipment by type to make that decision.
%However if we can make good predictions about the final quantity that is likely to arrive the MDP process will 
%become more accurate.

%Most existing works in the revenue management
%literature often use the average volume based on the type of the  shipment. 

%\subsection{Contributions}
\begin{packed_enum}
\item We introduce a unique data cleaning module using the concept of disguised missing values (DMVs). 
%An airline receives cargo booking data from agents who have different devices for volume measurements as well as collecting and transmitting data.
%For example, it is not uncommon to send shipment data to an airline via SMS, email, or over the phone.
The detection of DMVs significantly improves the overall quality of the prediction. (Section~\ref{section:dmv})

\item %As the actual volume are not known exactly at booking time, 
We propose to use Gradient Boosting Machines (GBMs) to build a volume prediction model that reduces the bias and the variance at the flight level.
Our model takes into account whether a booking might contain a DMV or not to increase the quality of the prediction.
%%We propose a volume prediction model that tightly integrates with the above decision making process and demonstrate the impact of prediction on the final revenue generation and offloading costs.
%We propose a volume prediction model that takes into account whether a booking might contain a DMV or not.
%We demonstrate the impact of prediction on the final revenue generation and offloading costs. 
(Section~\ref{section:predict})
%This model allows us to predict what the actual volume will be based on the initial booked volume.
%We demonstrate that using a prediction model for the final received volume can significantly lower the offloading cost and thus increase the profits.
%To the best our knowledge, this is the first published work that conclusively demonstrates that tightly integrating predictions into the decision making process improve overall efficiency of the system.

\item We model air-cargo revenue management as a prediction-driven sequential and stochastic optimization problem, where the state (total volume) is realized at departure time.
This model tightly integrates the above prediction model in order to make more reliable decisions.
%We demonstrate that our decision-making technique helps the airline company to increase the total revenue and decrease the offloading cost. 
(Section~\ref{section:decide})
%At each stage the system will make a decision on whether to accept or reject a cargo shipment for a particular flight in order to maximize the expected revenue for that flight.
%If a flight gets overbooked then the shipment is offloaded and that incurs a cost.
%When the MDP is making a decision the actual weight and volume is not known exactly.

\item We evaluate the AI-CARGO using both synthetic and real data taken from the airline company. We demonstrate that our prediction-based decision-making technique helps the airline company to increase the total revenue and decrease the offloading cost. (Section~\ref{section:results})
\end{packed_enum}

%We demonstrate the use of the AI-CARGO using both synthetic and real datasets in Section~\ref{section:results}.
We conclude this paper with a discussion, some lessons learned from the project, and future work in Section~\ref{section:discussion}.

%The rest of the paper is structured as follows. In Section two we briefly review the literature including
%works on relevant air-cargo revenue management from the operations research community.  In Section 3 we
%give an overview of the Artificial-Intelligence (AI) driven cargo System. In Section 4 we describe in detail
%the technical aspects of the system including each of the three modules. We give a simple but concrete example of 
%solving the stochastic dynamic program.
%In Section 5 we demonstrate the use of the AI-CARGO
%using both synthetic and real data. We conclude in Section 6 with discussion, learnings from the project and
%possible directions for future work.

\section{Related Work}
\label{section:relwork}
To the best of our knowledge AI-CARGO is the first known cargo revenue management
system which combines machine learning and mathematical optimization. We overview
related work in these subareas.

The discipline of Revenue Management (RM) is an advanced and well developed topic within the Operations Research (OR) community.  It has roots in the airline industry and has now expanded in other areas, including hotels and tourism~\cite{chiang}. The initial
focus in the airline industry was primarily on passenger RM, in particular how to price passenger seats
in order to maximize revenue~\cite{McGill:1999:RMR:767713.768451}. 
% TODO: this has to be moved to intro
%Passenger seats on a given flight as well as air cargo capacity can be seen as perishable supply - once the plane has flown the value of an unused passenger seat or cargo capacity becomes zero.
A strand of relevant work that has appeared in the data mining literature  is the problem of determining overbooking rate, forecasting the number of no-shows per flight, i.e.,~the percentage of bookings that were made but did not show up by departure time~\cite{Lawrence:2003:PPM:956750.956796,Hueglin:2001:DMT:502512.502578}. No-shows are a common problem in air-cargo too and several works have 
proposed solutions~\cite{lan2011regret,Popescu:20063}.

%Overbooking is a common practice for air-cargo and
%several recent papers have proposed solutions to estimate the overbooking rate

While RM for passenger seats is now a well developed area, the same is not true for cargo management. The first research overview
of issues surrounding air cargo revenue management were introduced in~\cite{kasilingam1997air}. Since then
the research literature has seen a steady growth with works including~\cite{popescu2006air,Budiarto_2018}. 
Our work closely follows the decision making paradigm introduced by ~\cite{amaruchkul2007single} for modeling both volume
and weight aspects of bookings in air-cargo RM. Our innovation is that 
%Our decision making approach is based on
%this paper except that we focus exclusively on modeling volume, as that is what
%our partner requested and that 
we integrate machine learning prediction into the decision making process.  
%A more recent survey on air cargo revenue management
%by Budiarto et al.~\cite{Budiarto_2018} provides an overview of the different techniques used in the OR community.

Decision making with machine learning can also fall under the umbrella of reinforcement
learning (RL)~\cite{sutton2018reinforcement}. However, much of the focus in RL
is on model-free approaches and on using function approximation for overcoming the curse of dimensionality phenomenon that is ubiquitous in sequential learning paradigms~\cite{sutton2018reinforcement,powell2007approximate,bertsekas2005dynamic}.
%  i.e., when the state transition function is
%not known.  In our case we do have a model of state transition,
%it is just that the booking values are only realized at the end of the booking
%horizon. RL uses function approximation (and thus machine learning prediction)
%to overcome the curse of dimensionality. In our case we use volume aggregation
%to reduce the size of the state space but use prediction to estimate
%the received volume (rcsvol) when the booking is first made. Our approach is
%also different from partially observable markov decision processes (POMDPs), in 
%that the state is fully realized at the end of the booking horizon~\cite{sutton2018reinforcement,powell2007approximate,bertsekas2005dynamic}.

The air cargo data that airlines collect is replete with errors because of the nature of the business. Shippers coordinate
with freight forwarders and airlines using various means including email, phone calls, SMS and this increases the possibility of data errors. 
Data cleaning, data curation and preparation is a very developed area within the database community~\cite{raman2001potter,elmagarmid2007duplicate} and there is tremendous potential
to put  these techniques into practice in the air cargo industry.
In order to integrate fine-grained predictive modeling we had to resort to various data cleaning approaches. 
A particular data quality problem we dealt with is Disguised Missing Values (DMVs)~\cite{Pearson:2006:PDM:1147234.1147247,Hua:2007:CDM:1281192.1281294,qahtan2018fahes}, 
where users instead of providing a NaN for unknown values use arbitrary yet valid data values.
The difference in our case is that DMVs are contextualized and conditioned, i.e., we only want to measure the impact of DMVs on the 
prediction of the final received volume.
%where users instead of providing a NaN for unknown values they use arbitrary valid data values.

\iffalse
Making prediction from structural data is a very common supervised task in data mining. In this context, the gradient boosting \cite{friedman2001greedy} technique has empirically proven itself to be highly effective for a vast range of supervised learning problems, currently achieving state-of-the-art performances in various tasks, from the prediction of booking demand \cite{ye2018customized} 
to medical application~\cite{rgbm2018}. 
In particular, extreme gradient boosting \cite{xgboost}, which supports several objective functions including regression, classification and ranking, has accumulated an impressive track record of winning machine learning competitions, such as Kaggle and KDDCup, gaining popularity in many applied ML works \cite{nielsen2016tree}.
\fi

\begin{figure*}[t]
\centering
\includegraphics[width=1\textwidth]{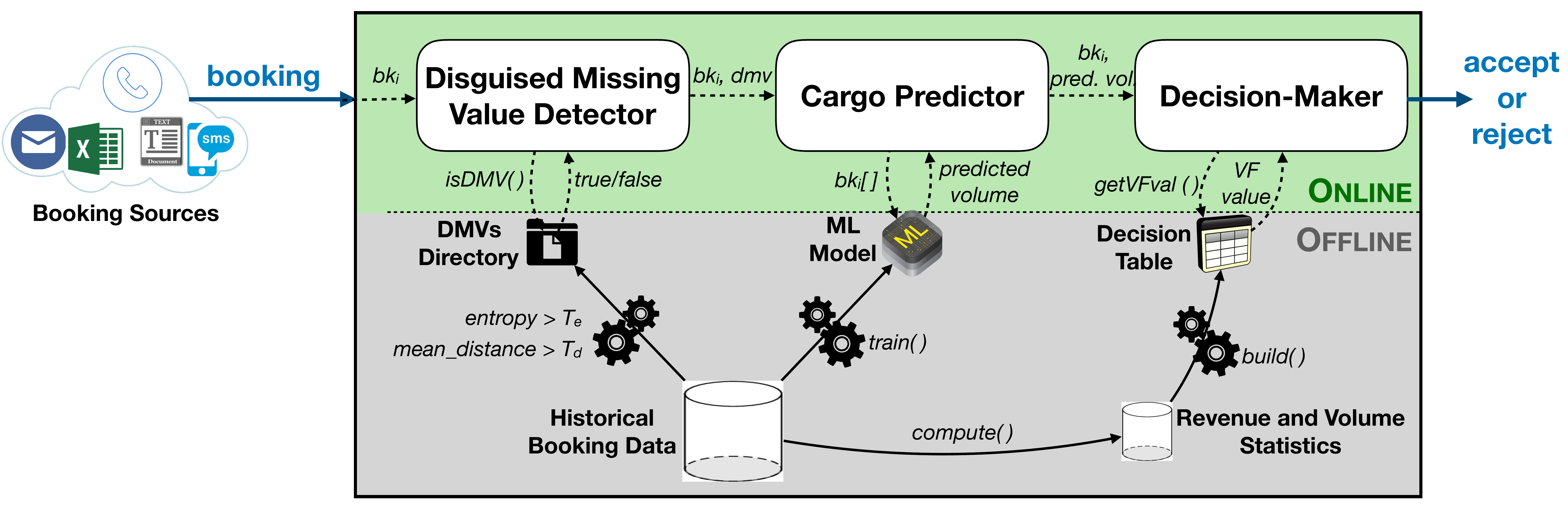}
\caption{The AI-CARGO system consists of three modules: (i) a disguised
missing value detector, a prediction component to estimate $rcsvol$ 
and a decision-making rule to decide which bookings to accept in order to 
maximize expected revenue.}
\label{aisystem}
\end{figure*}

\section{Proposed System: AI-CARGO}
\label{section:system}
AI-CARGO has three components as shown in Figure~\ref{aisystem}:
the (i)~Disguised Missing Value (DMV) Detector, (ii)~Cargo Predictor, and (iii)~Decision-Maker.
Given an incoming booking, the DMV Detector checks whether the booking has a DMV or not by consulting a directory of possible DMVs.
Then, the Cargo Predictor gets as input the booking from the DMV Detector (i.e.,~the booking with a DMV indicator), extracts the relevant features from the booking (including the DMV indicator), and predicts the volume expected to be tendered by the customer.
At the end of the pipeline, the Decision-Maker gets the incoming booking with the predicted volume that will be tendered and takes a decision whether to accept or not the booking via a stochastic dynamic program.
We detail each of these steps in the following three sections. 
%Table~\ref{tab:glossary} shows the terms we will be using in the following.
%prediction model which estimates rcsvol from bkvol 
%and other booking record features. The output of the DMV detector is used as feature in the prediction model (iii) a decision-making module which will use the
%prediction estimate of rcsvol to  accept or reject an incoming booking in order to maximize expected
%revenue. 

%\begin{table}[t]
%{\small
%\caption{Glossary of Terms}\label{tab:glossary}
%\begin{tabular}{|l|l|} \hline
%{\bf Terminology} & {\bf Definition} \\ \hline \hline
%bkvol (bkwt) & booked volume (weight) by shipping agent \\ \hline
%rcsvol (rcswt) & received volume (weight) by departure time \\ \hline
%chargeable weight & an IATA formula to convert vol. to equivalent wgt. \\ \hline
%DMV & Disguised Missing Value: a proxy for NaN \\ \hline
%D1V & a decision-rule for vector state $\vect{x}$\\ \hline
%D1S & a decision-rule for scalar state $x$\\ \hline
%D2V & a D1V that uses prediction for bkvol\\ \hline
%D2S & a D1S that uses prediction bkvol \\ \hline
%\end{tabular}
%}
%\end{table}

\subsection{Detecting DMVs}
\label{section:dmv}
\iffalse
\begin{table*}[t]
\begin{tabular}{|l|l|l|l|l|r|} \hline \hline
BOOKING ID & ORIGIN & DEST & PRODUCT & BOOKDED VALUE (BKVOL) & RECEIVED VOL (RCSVOL) \\ \hline
\hline
1201 & SIN & DEL & A & 10.23 & 18.45 \\ \hline
\vdots & \vdots & \vdots & \vdots & \vdots & \vdots \\ \hline
1503 & NYC & SYD & B & 10.23 & 5.05 \\ \hline
\end{tabular}
\caption{Example of Disguised Missing Value (DMV). 10.23 is a DMV. DMVs occur frequently
but not all frequenty occuring values are not DMVs.}
\end{table*}
\fi
The unique nature of the entire air-cargo ecosystem results in several practices that make bookings' values, e.g.,~booked volume ($bkvol$), not very reliable.
For instance, customers send booking information via SMSs, emails, text files, or even phone calls, which also causes a natural lag between the time a customer creates the booking and the time the booking is actually reflected in the central cargo system.

An important observation we made is that customers often send arbitrary but fixed values as proxies for NaN.
That is, when a customer first books an item the exact volume (or weight) that will be delivered is often not known.
However, instead of setting $bkvol$ to NaN, customers will often choose a fixed but arbitrary number, which will have no bearing on the received volume $rcsvol$. 
%Often these are not integer values.
In the data cleaning literature, these proxy values are often called Disguised Missing Values (DMVs)~\cite{Pearson:2006:PDM:1147234.1147247,qahtan2018fahes}. 
For example, consider a set of six bookings whose $bkvol$ was $10.23$ and $rcsvol$ were all different: $5.1$, $2.8$, $13.3$, $26.4$, $26.4$, and $2.8$.
Unless other features can explain the diverse range of values taken by $rcsvol$, $10.23$ is very likely to be a DMV.

Detection of DMVs is important as they can have a substantial impact on the prediction model.
To see how a DMV can effect prediction, consider the case of a linear regression model in a single dimension.
Suppose a linear model  $y = wx$ is learnt from examples
$(x_{i},y_{i})_{i=1}^{n}$.
Then, it is well known that 
\[
w = \frac{\sum\limits_{i=1}^{n}x_{i}y_{i}}{\sum\limits_{i=1}^{n}x_{i}^{2}}
\]
Now suppose $x_{dmv}$ and a associated  set of values $\{y_{1},y_{2},\ldots,y_{m}\}$ are
added to the training set. Let $w_{new}$ be the new updated parameter, which will be:
\begin{align*}
w_{new} & = \frac{\sum\limits_{i=1}^{n}x_{i}y_{i} + x_{dmv}\sum\limits_{j=1}^{m}y_{j}}{\sum\limits_{i=1}^{n}x_{i}^{2} + mx_{dmv}^{2}} \\
& = \frac{w + \frac{x_{dmv}\sum\limits_{j=1}^{m}y_{j}}{\sum\limits_{i=1}^{n}x_{i}^{2}}} 
{1 + \frac{mx_{dmv}^{2}}{\sum\limits_{i=1}^{n}x_{i}^{2}}} \\
\end{align*}
Depending upon the value of $\sum_{i}^{m}y_{j}$, the model might or might not be impacted by $x_{dmv}$.
For instance, if $\sum_{j=1}^{m}y_{j} = mwx_{dmv}$, then substituting in the above equation shows that $w_{new} = w$ and the DMV has no impact on the model.
However, if $\frac{1}{m}\sum_{i=}y_{j}$ deviates significantly from the straight line $y = wx$, the impact of $x_{dmv}$ can clearly be large.

We thus use the mean distance and entropy information to compute two features for each $bkvol$ in order to detect DMVs.
For each distinct $bkvol$, denoted as $u_{i}$, let $V_{i} = \{v_{i,1},\ldots,v_{i,n_{i}}\}$ be the set of $rcsvol$s that appear in the data.
Then, we define $g_{1}$ to quantify how much the average of $rcsvol$s deviates from the associated $bkvol$.
Formally,
\begin{align*}
g_{1}(u_{i}) & = \left(\frac{1}{n_{i}}\sum\limits_{k=1}^{n_{i}}v_{i,k} - u_{i}\right)^{2}
%g_{2}(u_{2}) & = \frac{-\sum\limits_{k}p_{k}
\end{align*}
Then we redefine a second feature $g_{2}$ to capture the entropy of the set $V_{i}$.
The higher the entropy the more likely that $u_{i}$ is a DMV.
We normalize the entropy score in order to upperbound it at one.
If $V_{i}$ consists of $K$ distinct elements, then let $V_{i,k}$ be the the k-th bucket of $V_{i}$ and define $p_{k} = \frac{|V_{i,k}|}{|V_{i}|}$.
We thus define $g_{2}$ as
\begin{align*}
g_{2}(u_{i}) = \frac{-\sum\limits_{k=1}^{K}p_{k}\log p_{k}}{\log n_{i}}
\end{align*}

\begin{figure}[t!]
\includegraphics[width=.9\columnwidth]{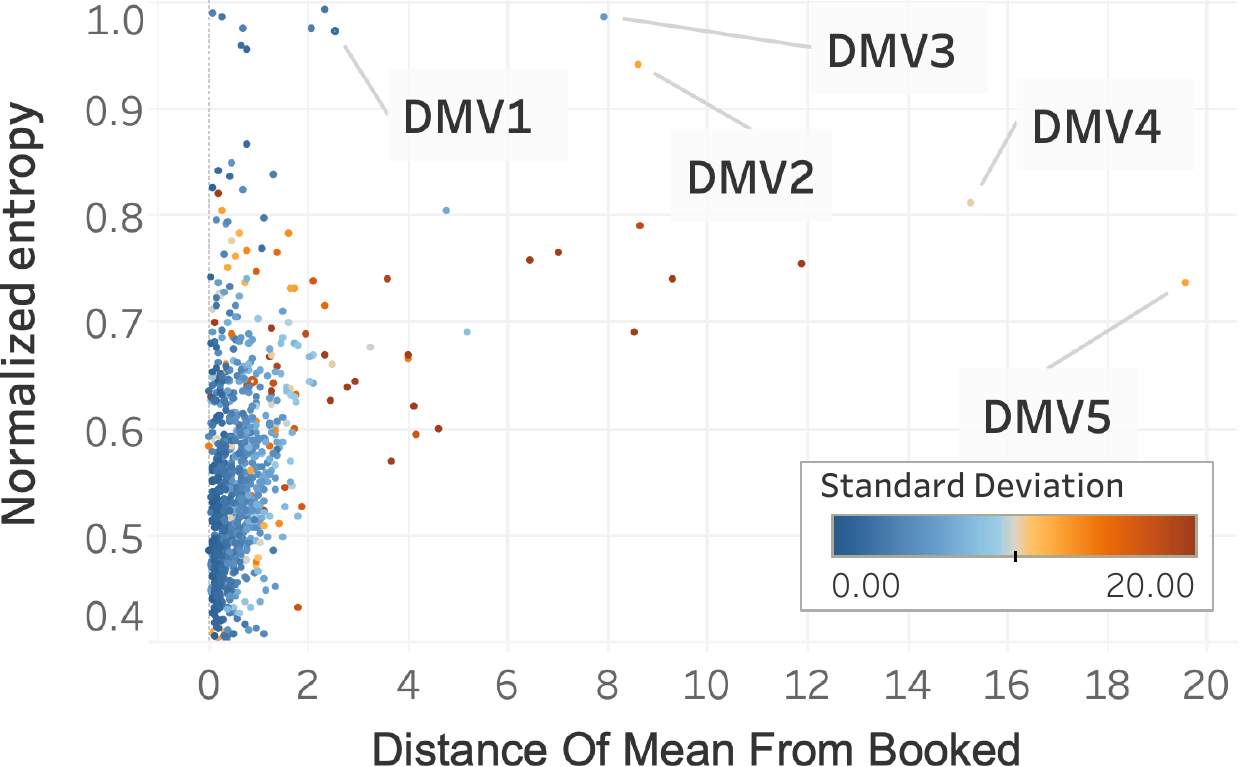}
\caption{DMVs identified by the DMV score method}
\label{dmv_result}
\end{figure}

When $g_{1}$ and $g_{2}$ are above some threshold it means that the $bkvol$ is a DMV.
Using features $g_{1}$ and $g_{2}$, we offline create the DMVs directory illustrated in the left-bottom part of Figure~\ref{aisystem}.
In more detail, from our historical booking data, we map every frequent\footnote{We used a threshold of 0.01\%.} distinct value into a two-dimensional feature space and visually browse the space to set a cutoff threshold.
We then denote every $bkvol$ value that crosses the threshold as DMV and put it into a DMVs directory.
Figure~\ref{dmv_result} shows an example of the two-dimensional DMV space, where the x-axis is $g_{1}$ and the y-axis is $g_{2}$, and highlights several known examples of DMVs, confirmed by domain experts.
Our goal is to let the algorithm learn that if a booking contains a $bkvol$ that is likely to be a DMV, the, it should deemphasizes the $bkvol$ in the prediction of $rcsvol$.
This approach also makes it easier to handle DMVs in a production environment. 

\iffalse
To begin the DMV process we first extracted the bkvol values that appeared
in  0.01\% of the bookings or more. For each of the extracted bkvol values, we
built the set of unique rcsvol in the dataset and computed the probability of each rcsvol given the bkvol, in order to calculate the normalized entropy of the rcsvol set. We also computed the mean rcsvol in the dataset for each bkvol and its distance from the bkvol.
\fi

\begin{figure*}[t!]
\centering
\includegraphics[width=0.8\textwidth]{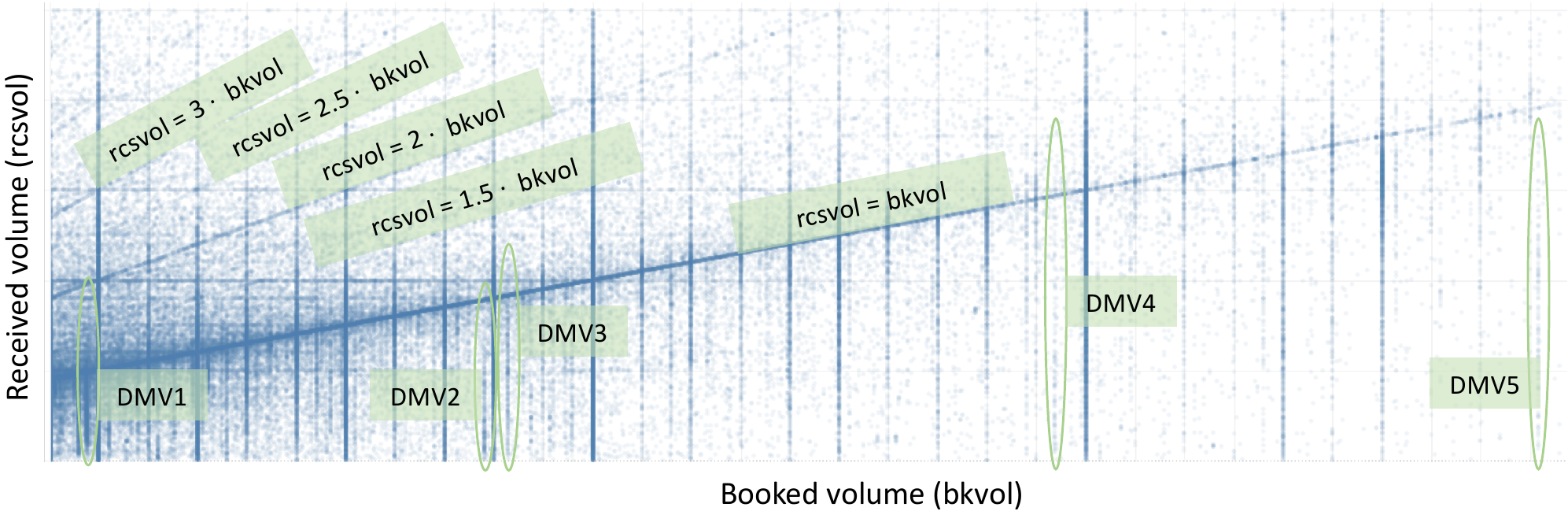}
\caption{Values of bkvol and related rcsvol in shipments from historical data for a given range of bkvol. Circled in green are vertical patterns that are evidently disguised missing values.}
\label{fig:dataplot}
\end{figure*}

\subsection{Predicting Cargo Volume}
\label{section:predict}
Given an incoming booking with its DMV flag, AI-CARGO proceeds to predict $rcsvol$ (i.e.,~the received volume by departure time) for the given booking, as shown in Figure~\ref{aisystem}.
To do so we need to offline build a model using historical data as illustrated in the middle-bottom of Figure~\ref{aisystem}. We thus need to (i)~decide on the features and~(ii)~decide on the algorithm to use.
Predicting $rcsvol$ is quite challenging as the $bkvol$ is usually quite different from $rcsvol$.
Figure~\ref{fig:dataplot} illustrates this difficulty.
The vertical lines provide a clear indication that we need other features besides $bkvol$ to have any chance of accurately predicting $rcsvol$.

We thus experimented with extracting different feature combinations until we settled on a set that provides a good compromise between model complexity and accuracy.
More formally, given a sample of bookings, we formed a feature set $\vect{X}$ and mapped each booking $i$ as an element $\vect{x}_{i} \in \vect{X}$ and the $rcsvol$ as $y_{i} \in \vect{R}^{+}$.
The prediction task then becomes a regression problem, where we have to learn a function 
\[f_{\theta}: \vect{X} \rightarrow \vect{R}^{+} \]

%For each bkvol, the entropy and variance of its rcsvol is very high. This is true not 
%only for disguised missing values but for any bkvol, and it is made particularly clear by looking at the vertical lines in very frequent bkvol.

\begin{figure}[t!]
\includegraphics[width=0.7\columnwidth]{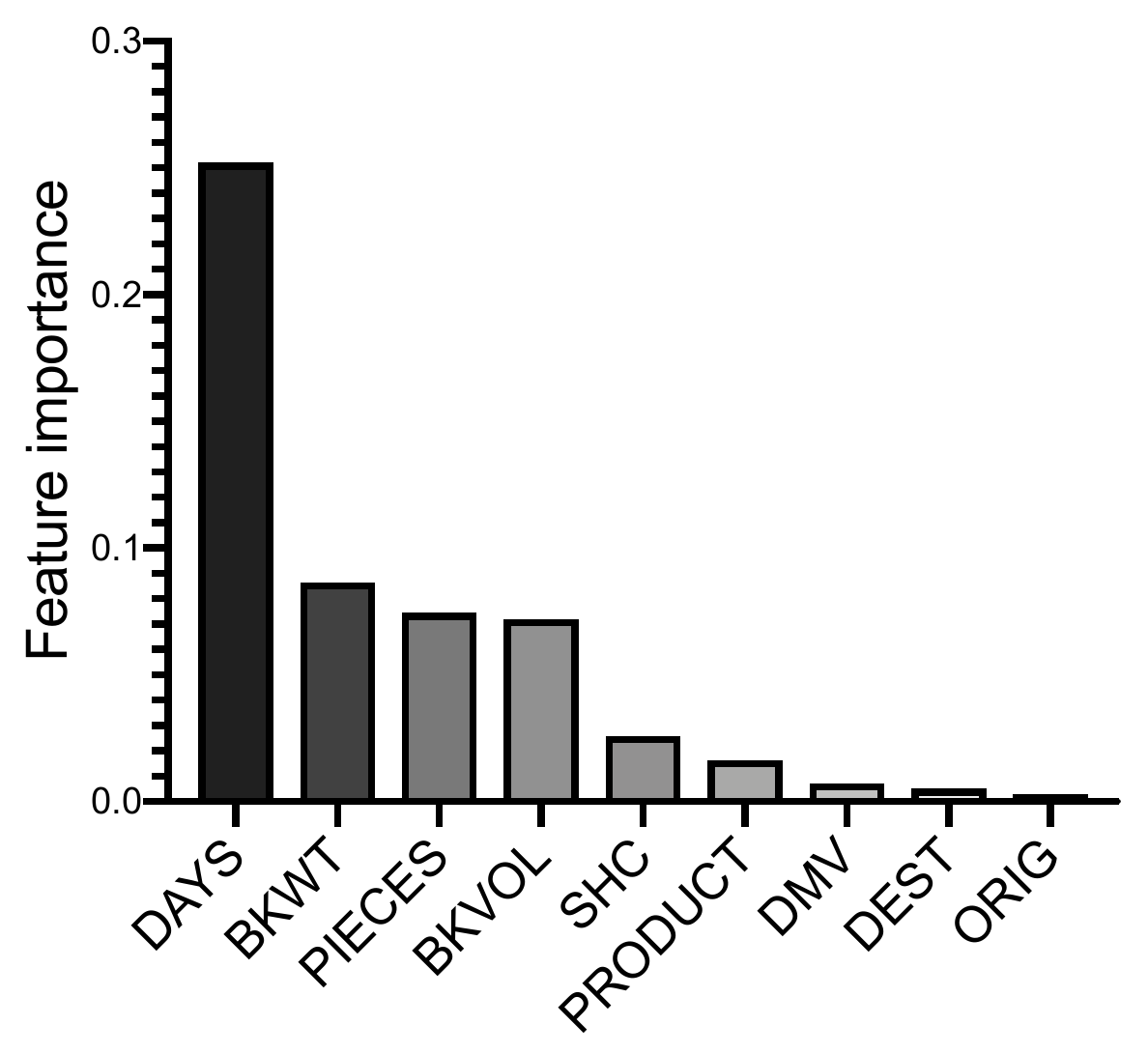}
\vspace{-0.2cm}
\caption{Features importance. For categorical features, such as product, the value of the category with the maximum importance is shown.}
\label{figure:features_importance}
\end{figure}

Figure~\ref{figure:features_importance} shows the set of features we use, which were the most important ones, for predicting $rcsvol$:
\begin{itemize}
\item \textbf{Days until departure (DAYS):} The most important feature by far is the number of days between the booking time and the departure time.
Bookings closer to the departure time tend to be more accurate.
In fact, bookings that are time-stamped several days before departure day tend to show a clear pattern of overbooking from the customer side (and hence of using DMVs). 
It is natural for shippers to overbook as in the air-cargo business there is no penalty for that.

\item \textbf{Booked weight (BKWT):} Contrary to $bkvol$, which usually tends to be a DMV, $bkwt$ is a valuable information.
This is because shipping agents have a much more accurate information of $bkwt$ as they have access to high quality weighing machines.
Indeed, instruments for accurately measuring volume are not that widespread~\cite{slager,beidermand2002new}.
Thus, being easier to measure, $bkwt$ is on average more precise. 
% and the model can learn the density given the product type and sufficient number of samples.

\item \textbf{Number of pieces (PIECES):} A shipment may consist of a number of equal units.
Diagonal lines in Figure~\ref{fig:dataplot} suggest that bookings frequently differ from the tendered shipments in the number of pieces rather than in their volume.
Thus, knowing the number of pieces is useful in predicting possible outcomes at receiving time.
For example, if two pieces where booked for a $bkvol$ of $12m^3$, with volume for each piece of $6m^3$, it is unlikely that a single piece will be split and the $rcsvol$ will become $4m^3$.
It is in fact much more likely that it may become $6m^3$, $18m^3$, or $24m^3$.

\item \textbf{Booked volume (BKVOL):} We observed that, despite of DMVs, the booked volume, $bkvol$, is still an important feature for predicting $rcsvol$.
This is because when $bkvol$ is not a DMV, it tends to be precise.
%The booked volume, $bkvol$, is by far the most important feature for predicting the $rcsvol$, despite the fact that many bkvols are actually DMVs.

\item \textbf{Shipment code (SHC):} This is a set of codes to instruct how the shipment must be handled, e.g.,~live animals or perishables.
This feature ended up being important as it specifies over (or complements the) the product type explained here below.
We encode the shipment code feature as a binary vector with one element for each shipment code (one-hot encoding).

\item \textbf{Product type (PRODUCT):} We observed that the patterns in $rcsvol$ vary with different product type.
In theory product type should be a highly informative feature, but we observed that the distribution of product types is skewed.
%Infact the distribution of product types is highly s%reliability of a booking varies with the product type. [todo: ref to image?]

\item \textbf{DMV Flag (DMV):} Because DMVs are frequent and must be dealt within a production environment, we decided not to remove DMV data from the training set.
At the same time, giving their negative impact in the prediction, it is important to know if a booking has a DMV for $bkvol$.
% as they exhibit a different behavior compared to other shipments.
For this reason, we provide a flag, which is obtained by the DMV Detector based on historical data.

\item \textbf{Destination (DEST):} We also consider the destination as a feature, even if the destination alone is a weak predictor for $rcsvol$.
This is because in conjunction with product type it becomes possible to elicit subtypes within products and thus reduce the variance.
%For example the product type maybe general but in conjunction with destination
%type it becomes possible to elicit subtypes within products and thus reduce the variance.
%airport of the shipment. Late changes in shipment volumes may depend on changes of the other end's demand.

\item \textbf{Origin (ORIG):} The origin airport of the shipment. This feature allows capturing the average behavior of booking agents from each location. 

 %\item 
 %is frequently a multiple or a unit fraction of the bkvol.

%\item \textbf{Agent behavior:} We quantified agent behavior in terms of their history
%to be accurate and on time. For example, for each agent we calculated the percentage of
%times they were late in delivering the shipment and whether they were more accurate 
%for certain product types or certain destinations.
\end{itemize}

%The feature vector has one element for each numerical variable, such as the bkvol value, and $k$ elements for each categorical variable, such as the product type, where $k$ is the number of unique values that the variable may assume in our dataset, following the one-hot encoding representation.

%\noindent
%{\bf Gradient Boosting vs. Random Forests:}
We also experimented with using both random forests (RFs) and gradient boosting machines (GBMs)~\cite{xgboost} for building the model.
GBMs are ensemble methods and are known to perform well ``out of the box''.
They can also easily handle a mixture of datatypes including numeric and categorical data.
%Deep Learning methods are usually not competitive in tabular settings with mixed datatypes
%\todo{citation}.
Recall that for a prediction problem the error can be decomposed into a sum of bias and variance~\cite{tibshirani}.
On the one hand, RFs reduce the error by reducing variance as they combine independently generated deep trees on bootstrapped samples.
GBMs, on the other hand, reduce the bias by building shallow trees in a sequential manner, where each subsequent tree is trained by using the dependent variable as the residuals of the previous one.
In our case, even though we trained the model to make predictions at the booking level, we were primarily interested in making flight level predictions, which are an aggregation of booking level predictions.
Therefore, while GBM predictions fluctuate more and individual predictions are further from the actual value, the differences cancel each other out at the flight level, that is, the aggregation of bookings at the flight level will automatically result in variance reduction.
This has been confirmed by evaluating both models for booking level and flight level prediction. On the booking level, the variance in GBM predictions is more than 5 times higher than RF predictions. However, at the flight level, the mean absolute error of RF is 87.1\% higher than the GBM error.  
%To get a better understanding of how GBMs and RFs may behave, consider a synthetic example shown in Table~\ref{bias}.
%A flight consists of four bookings (first row) and the actual $rcsvol$s are given in the second row.
%The flight level $rcsvol$ is the sum of booking level $rcsvol$s.
%The GBM and the RF predictions are in row three and four.
%While GBM predictions fluctuate more and individual predictions are further from the actual value, the differences cancel each other out at the flight level.
%This is because GBMs reduce error by reducing bias.
%RFs, on the other side, have smaller individual error, but they tend to be biased when aggregated at the flight level.
For these reasons, in our production deployment we have used GBMs.
%, but it will be an informative exercise to deploy RFs too and compare the two in a real setting.

%\begin{table}[t!]
%\caption{Synthetic example to illustrate difference between GBMs and RF. The errors
%in GBM cancel out more at the aggregate flight level.}
%\label{bias}
%\vspace{-0.2cm}
%\begin{tabular}{|l|l|l|l|l|l|}  \hline \hline
%& bk1 & bk2 & bk3 & bk4 & flt-level \\ \hline
%actual & 10 & 15 & 12 & 18  & 55 \\ \hline
%GBM predictions & 6 &  20  & 10  & 20 & 56 \\ \hline
%RF predictions & 11 & 16 & 14 & 17 & 58 \\ \hline
%\end{tabular} 
%\vspace{-0.5cm}
%\end{table}

\iffalse
We apply a Gradient Boosting Machine, namely XGBoost, to learn a non-linear regression model predicting the rcsvol give the above features of each shipment. We choose GBM for its state-of-the-art accuracy, generalization and explainability. A grid-search is performed over several combinations of hyper-parameter, in a 3-fold cross-validation, with the aim of maximizing the prediction accuracy while minimizing the overfitting.
\fi

\subsection{Decision Making}
\label{section:decide}
Once a booking is DMV-tagged and its $bkvol$ is predicted, the Decision-Maker creates an acceptance/rejection suggestion for the given booking.
Before digging in how it does so, let us first state what the problem of decision making in the context of airline cargo booking is.
For any flight, capacity is a perishable quantity, i.e.,~once the flight takes-off the capacity is lost.
Therefore, an airline wants to accept bookings that will maximize revenue.
The problem can be seen as a generalization of the classic Knapsack problem with two caveats:
(i)~cargo bookings appear over time and the exact volume (weight) of the shipment becomes available only at departure time.
We, thus, model this problem as a stochastic dynamic program~\cite{amaruchkul2007single,powell2007approximate}.

We start by defining a state vector $\vect{x} = (x_{1},\ldots,x_{m})$.
Each $x_{i}$ is the number of items of type $i$ assigned to a flight.
A type is a pre-defined category, like fresh food or pharma.
The state $\vect{x}$ evolves with time $t$.
We define the value function $VF(x,t)$ as the expected revenue from the flight given that at time $t$ the flight is in state $\vect{x}$.
We label departure day as time $t=0$ and the booking horizon extends up to time $t=T$.
Thus, time flows backwards.
We model a single flight whose volume capacity $k_v$\footnote{For non-cargo flights $k_v$ varies depending upon passenger load.} is fixed and known.
We also discretize time and in each time bin $t$ the probability of an item $i$ being received for a booking is $p_{i,t}$.
We assume that at each time step only one shipment can arrive for booking.
We define $p_{0,t}$ as $ 1  - \sum_{i=1}^{m}p(i,t)$ as the probability that no booking will show up in time period $t$.
In practice, when an agent books an item of type $i$, it is accompanied by a booked volume $bkvol_i$.
When the item finally arrives for shipment the received volume is $rcsvol_i$.
The revenue received from the item $i$ is $R(rcsvol_{i})$, where $R()$ is typically an increasing and concave function of volume.
Recall that during booking time the airline only knows $bkvol_{i}$ and not $rcsvol_{i}$.
Thus, it is common to make a decision about whether to accept or reject a booking based on the average volume of type $i$, $\bar{v_{i}}$. 

We can now define the value function $VF(\vect{x},t)$ as a recursive function (Bellman's Equation) in order to maximize the overall expected revenue~\cite{amaruchkul2007single}:
\begin{align*}
VF(\vect{x},t) & = \sum_{i=1}^{m}p_{i,t}\max\{R(\bar{v}_{i}) + VF(\vect{x + e_{i}},t-1), VF(\vect{x},t-1)\}&\\
 &\quad + p_{0,t}VF(\vect(\vect{x},t-1)), \quad t = 1,2,\ldots, T \\
VF(\vect{x},0) & = -h_{v}\left[\sum_{i=1}^{m}{x_{i}}\bar{v}_{i} -k_v\right]^{+}
\end{align*}
where $[a]^{+} = \max\{a,0\}$.
We now explain the above recursive equation.
When the state is $\vect{x}$ at a given time $t$ then $VF(\vect{x},t)$ is the expected revenue over the full time horizon of the booking.
At time step $t$, the probability of a shipment of type $i$ arriving is $p_{i,t}$.
If the booking is accepted then the state will transition to $\vect{x + e_{i}}$, where $\vect{e}_{i}$ is the one-hot binary vector with a 1 at the $i$-th location.
By accepting the booking, the expected revenue will be $R(\bar{v}_{i})$. 
However, the booking of item $i$ will only be accepted if the revenue $R(\bar{v}_{i}) + VF(\vect{x + e_{i}},t-1)$ is greater than not accepting the booking and transitioning one step towards departure while staying in the same state, i.e.,~$VF(\vect{x},t-1)$.
At time $t=0$ and in state $\vect{x}$, the $VF(\vect{x},0)$ captures the cost of off-loading, which is proportional ($h_v$) to the total expected volume $\sum_{i}x_{i}\bar{v}_{i}$ minus the capacity $k_v$.
For example, if the expected volume is $100$ units and the capacity $k_v$ is $50$, then the off-loading cost is $-50h_{v}$. 

Having defined the value function $VF(\vect{x},t)$, the decision rule (D1V) at each time step $t$, which determines whether to accept or reject an incoming shipment of type $i$ is given as:
\begin{equation*}
\boxed{
\begin{array}{ll}
D1V: & R(\bar{v}_{i}) + VF(\vect{x + e_{i}},t-1) >  VF(\vect{x},t-1)
\end{array}
}
\end{equation*}

However, we can integrate prediction in the decision-making by modifying the decision rule.
For example, suppose our predictive function is $f_{\theta}$ (as defined in Section 3.2\footnote{We have overloaded the $f_\theta$ signature to emphasize the role of $bkvol$}), i.e.,~given a booked volume $bkvol_{i}$ of type $i$, $f_{\theta}(bkvol_{i})$ is the predicted received volume ($\hat{rcsvol_{i}}$).
We, then, have a new decision rule (D2V) as 
\begin{equation*}
\boxed{
\begin{array}{ll}
%D2V: &  R(f_\theta(bv_{i})) + VF(\vect{x + e_{i}},t-1) >  VF(\vect{x},t-1)
D2V: &  R(f_\theta(bkvol_{i})) + VF(\vect{x + e_{i}},t-1) >  VF(\vect{x},t-1)
\end{array}
}
\end{equation*}

\noindent
{\bf The Curse of Dimensionality:}
It is worth noting that the construction of $VF(\vect{x},t)$ suffers from the well-known curse of dimensionality of dynamic programming~\cite{powell2007approximate,bertsekas2005dynamic}.
For example, suppose there are $m$ items and the number of time periods is $T$.
Then, the size of the state space is exponential in $m$\footnote{It is $S(T,m)$, Stirling number of second kind}.
An approximate solution to escape the exponential blow-up is to use aggregate $x = \sum_{i}x_{i}\bar{v_{i}}$.
This makes the state space one-dimensional scalar-valued, instead of vector-valued, of maximum size.
This state space is bounded by $M\times T$, where $M$ is the maximum possible volume booked for any type.
The construction of $VF(x,t)$ becomes considerably simplified and the decision rule D2V then becomes D2S:
\begin{equation*}
\boxed{
\begin{array}{ll}
%D2S: &  R(f_\theta(bv_{i})) + VF(x + f_\theta(bv_{i}),t-1) >  VF(x,t-1)
D2S: &  R(f_\theta(bkvol_{i})) + VF(x + f_\theta(bkvol_{i}),t-1) >  VF(x,t-1)
\end{array}
}
\end{equation*}

\iffalse
A weakness of the model is that it is only weakly data-driven. For example, at time $t$ the model does not take into
account the information received between $T$ and $T - t - 1$ as it only designed to work with averages. 

To account for a more real time setting we have designed the following model.
Suppose we are at time step $s$ and the state of the model is $\vect{\underline{x_{s}}} = (\underline{x_1},\ldots,\underline{x_s})$.

\begin{align*}
VF(\vect{x},t,s) & = \sum_{i=1}^{m}p_{i,t}\max\{r(\hat{v}_{i,t}) + VF(\vect{x + x_{s} + e_{i}},t-1,s),& \\ 
VF(\vect{x}+&\vect{x_{s}},t-1)\}  + p_{0,t}VF(\vect(\vect{x+x_{s}},t-1,s)), t = 1,2,\ldots, T & \\
VF(\vect{x},0,s) & = -h_{v}(\sum_{i=1}^{m}\sum_{l=T}^{s+1}\hat{v}_{i,t} + E\left(\sum_{i=1}^{m}\sum_{k=1}^{x_{i}}V_{ik} \right)) &
\end{align*}

The decision rule to accept shipment $i$ when in state $\vect{x}_{t}$ is 
\[
r_{i}  + VF(\vect{x}_{t} + \vect{e}_{i} ,t-1,t) > VF(\vect{x}_{t},t-1,t)
\]
\fi
%!TEX root = main.tex

%\subsection{Illustrative Example}

\begin{figure}[t]
\centering
\includegraphics[width=0.3\textwidth]{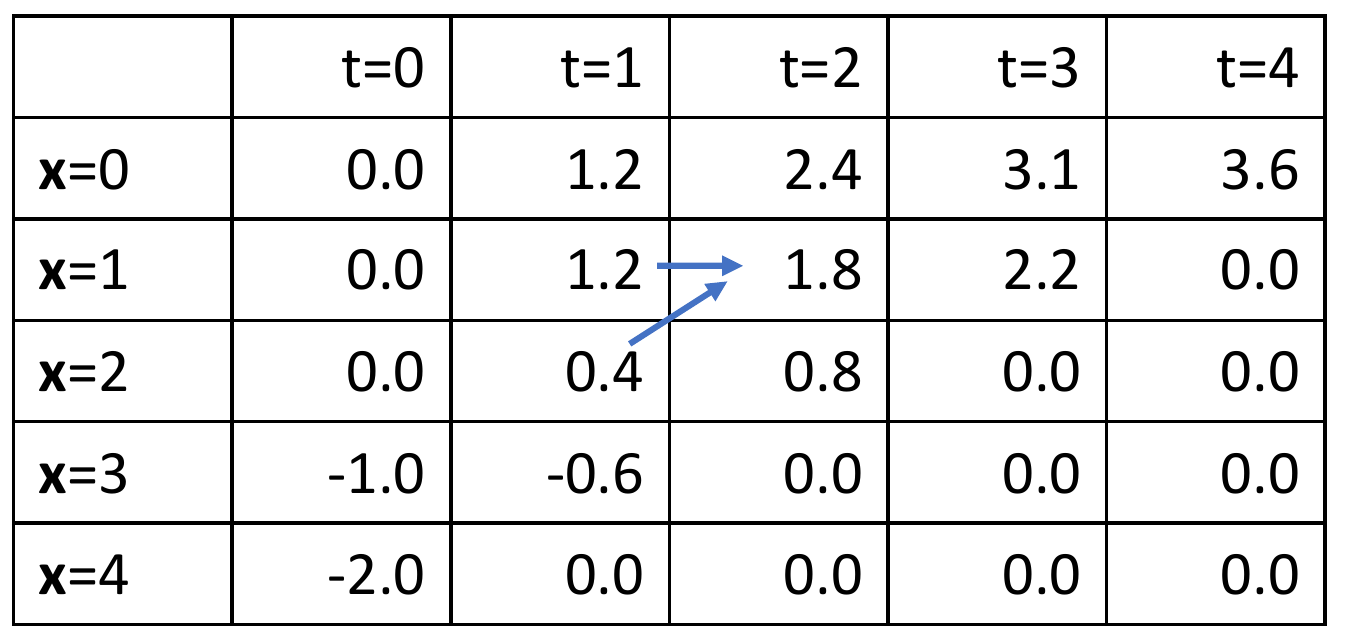}
\caption{The Value Function (VF) corresponding to Example 1.
Each cell corresponds to the revenue that
will be accrued starting from state $x$ at time $t$. Note the cell values are computed using backward induction
as time is going backwards.}
\label{mdpf}
\end{figure}

\noindent
{\bf Illustrative Example 1:}
To illustrate how dynamic programming is used to form the $VF$ function, we will work through a simple scenario shown in Table~\ref{mdpt}.

\begin{table}[h]
\caption{Sample cargo characteristics}
\label{mdpt}
\vspace{-0.2cm}
\centering
\begin{tabular}{|r||l|r|r|}  \hline
& & type1 & type2 \\ \hline
1 & volume & 1 & 1  \\ \hline
3 &revenue ($\rho$) & 1 & 2   \\ \hline
4 & prob. arrival in $t$  & 0.4 & 0.4   \\ \hline
5 & prob. no booking in $t$ &  \multicolumn{2}{|c|}{0.2}  \\ \hline
6 & max capacity ($k_v$) &  \multicolumn{2}{|c|}{2}  \\ \hline
\end{tabular}
\end{table}

\noindent We assume there are two shipment types: type1 and type2.
Both types can arrive for a booking with a probability of 0.4 in any time step and the probability that no shipment will arrive for any booking is 0.2.
The revenue for type1 is $1$ and for type2 is $2$, while the volume for both types is fixed at $1$ unit.
Recall that time is labeled in a reverse order, i.e.,~departure time is $0$ and booking horizon extends up to time $t=4$.
To compute the value function $VF$, we proceed backwards for each state $\vect{x}$.
Now the state is a two-dimensional vector $\vect{x} = (x_{1},x_{2})$, where $x_{1}$ and $x_{2}$ are  the number of bookings of type1 and type2 respectively.
However, we collapse $\vect{x}$ into $x = x_{1}\bar{v_{1}} + x_{2}\bar{v_{2}} = x_{1} + x_{2}$ as we have assume that the volume booked has a value of $1$.
The different values of $x$ are shown as rows in Figure~\ref{mdpf}.
We first have to populate the first column of Table~\ref{mdpt}.
For example, $VF(0,0) = -\max(0-k_v,0) = 0$ as $k_v = 2$ and $VF(3,0) = -\max(3-2,0) = -1$. 
As an example, we compute $VF(1,2)$.
\begin{align*}
VF(1,2) = & 0.4\max(1 + VF(2,1), VF(1,1)) \\
        & + 0.4*\max(2 + VF(2,1),VF(1,1) + 0.2*VF(1,1) \\
= & 0.4*(1 + 0.4) + 0.4*(2 + 0.4) + 0.2*1.2 \\
= & 1.76 \approx 1.8
\end{align*}

\iffalse
Let $\vect{x}$ be the three-dimensional state of cargo.
The terminal value function is
\[VF_{0}(\vect{x}) = -E\left[h\left(\sum_{i=1}^{3}\sum_{k=1}^{x_{i}}V_{ik}\right)\right] 
\]
where $h(\alpha) = c.(\alpha - k_{v})^{+} \equiv max\{c,0\}$  is the overbooked volume and $c$ is the margin cost of overbooking.
For this example $c=1$. 

Let $x = \sum_{i=1}^{3}x_{i}\bar{v}_{i}$ be the total expected volume of cargo. For example if $\vect{x} = (1,2,1)$ then
$x = 1.\bar{v}_{1} + 2.\bar{v}_{2} + 1.\bar{v}_{3} = 1.2 + 2.1 + 1.3  = 7$.
\fi

%!TEX root = main.tex

\section{Results}
\label{section:results}
We first evaluate our proposed AI-CARGO system in terms of revenue and costs.
We then show an in-depth analysis of our techniques.
Note that the prediction module of AI-CARGO has been deployed in a large international airline company\footnote{Note again that we cannot convey the name for confidentiality reasons.} and the results reported are from the production environment. 

\subsection{Dataset}

\begin{table*}[t]
	\caption{The top table shows the probabilities ($p_{i}$) of making a booking for a product type $i$. The bottom table shows the probabilities ($p_{t}$) of a booking arriving in a time period $t$.}
	\label{tbl:type_prob}
	\begin{tabular}{|c|llllllllll|}
		\toprule
		{\bf Product type} &Type1&Type2&Type3&Type4&Type5&Type6&Type7&Type8&Type9&Type10\\\midrule
		{\bf $p_{i}$} &  0.856 & 0.042 & 0.036 & 0.035 & 0.012 & 0.007 & 0.005 & 0.003 & 0.002 & 0.002
		\\\bottomrule
	\end{tabular}
	\quad
	\begin{tabular}{|c|llllll|}
		\toprule
		{\bf Time period}&1-10&11-20&21-30&31-40&41-50&51-60\\\midrule
		{\bf $p_{t}$} &  0.05 & 0.03 & 0.009 & 0.004 & 0.003 & 0.005
		\\\bottomrule
	\end{tabular}
\end{table*}

\iffalse
\begin{table}[h]
	\begin{tabular}{|cllllllllll|}
		%	\toprule
		time period	&1-10&11-20&21-30&31-40&41-50&51-60\\\midrule
		$p_{t}$ &  0.05 & 0.03 & 0.009 & 0.004 & 0.003 & 0.005
		\\\bottomrule
	\end{tabular}
	\caption{Probabilities of getting an incoming shipment for each of the 60 time steps in the booking horizon.}
	\label{tbl:time_prob}
\end{table}
\fi

\begin{figure*}[th!]
	\includegraphics[width=0.9\textwidth]{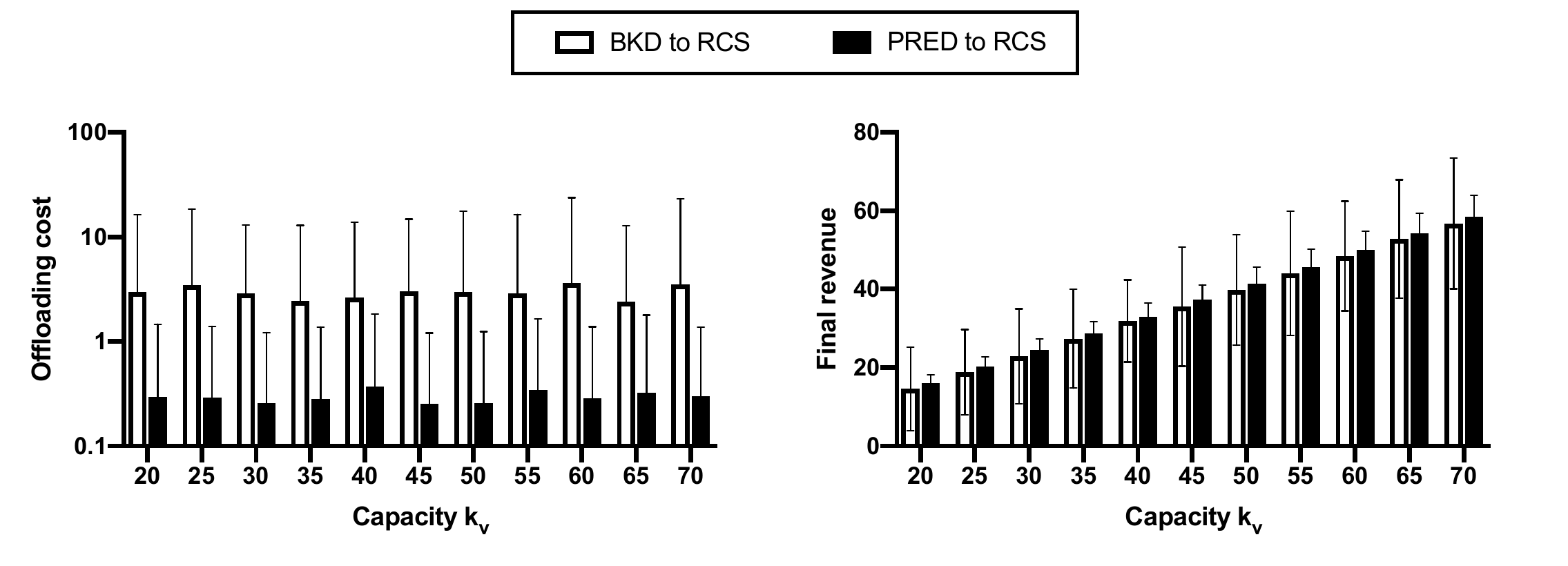}
	\caption{Offloading cost and the revenue generated for varying levels of capacity $k_v$}
	\label{fig:real_rev}
\end{figure*}

We obtained a real dataset spanning two years (June 2016- August 2018)  of booking records from the cargo IT team of the airline company.
Each booking record consists of several attributes including, booking date, origin, destination, agent, booking volume ($bkvol$), product type, received date, departure datetime, and received vol ($rcsvol$). 
We use this dataset to detect the DMVs and build the ML model for predicting $rcsvol$ using all other attributes.
As we do not have real information on the revenue and offload costs, we create simulated data as proposed in~\cite{amaruchkul2007single} to evaluate our decision making approach.
To create the simulations from the real dataset, we compute the probabilities of the product types from the real dataset: 
For each type $i$, we compute

\begin{align*}
p_{i} = \frac{\text{\# bookings with product $i$}}{\text{\# total bookings}}
\end{align*}

The related probabilities of the ten most frequent product types are shown in the top Table~\ref{tbl:type_prob}. We observe that the product type frequencies are skewed, with the most frequent product type with a probability of $0.856$. 
%This makes our prediction process more challenging.
Then, we split the booking horizon into 60 equal time steps and we compute the probability of a booking arriving at time step $t$ as

\begin{align*}
p_{t} = \frac{\text{\# shipments at time $t$}}{\text{\# shipments in dataset}}
\end{align*}

For the sake of simplicity and to replicate the same formulation as in~\cite{amaruchkul2007single}, we compute a single probability for 6 different intervals of time steps, resulting in 10 time steps per interval.
For each interval we take the average of the 10 single time steps that belong to this interval. The results are shown in the bottom Table~\ref{tbl:type_prob}.
Given $p_i$ the probability of an incoming type $i$  at any time, and $p_t$ the probability of getting any type of booking at time $t$, the probability of getting a booking of product type $i$ at time $t$ is $p_{i,t} = p_i p_t$.

%subsection{AI-CARGO Interface}
%Part of the AI-Cargo interface that is deployed is shown in Figure~\ref{charts}. 
%\begin{figure*}[t]
%%\includegraphics[width=0.9/textwidth]{figures/dashboard}
%\caption{The AI-Cargo interface which shows the changes in total rcsvol upto departure
%date. The green line is the prediction of the total volume that will be received for
%bookings made upto that time. The blue line is the actual received volume and the red
%line are the cancellations. Note how the green and blue lines converge. Also notice
%that the actual received volumes pick up few days before departure day.}
%\label{charts}
%\end{figure*}

\subsection{AI-CARGO Revenue Analysis}
We first evaluate our proposed AI-CARGO system in terms of revenue benefits and offload costs.
Using the data described above we simulate the whole AI-CARGO system pipeline including data cleaning, prediction, and decision-making as follows. First, we compute the $VF(x,t)$ table by using the entire dataset and taking the average booking volume for each time step.
Then, we consider two different test cases in order to evaluate the power of our system by combing predictive modeling with decision making:
\begin{packed_enum}
  \item \textit{BKD to RCS}: no prediction is made. The decision to whether accept or reject an incoming booking by applying $DS2$  is based on the reported
bkvol, i.e.,~$f_\theta(bv_i)=bv_i$. Final offloading cost is then calculated based on the rcsvol.
  \item \textit{PRED to RCS}: The received booking is processed for DMV identification and the resulting feature vector is used to output a prediction $f_\theta(bv_i)$ using our prediction model. The decision to accept or reject an incoming shipment using $D2S$ is based on this predicted volume.
\end{packed_enum}

At each time step we draw bookings from the dataset following the probabilities of Table~\ref{tbl:type_prob} and apply the decision rule $D2S$. 
Figure~\ref{fig:real_rev} shows the results on eleven different flight capacities $k_v$ and ten thousand flights each, for a total of 220 thousand flights. 

\iffalse
\begin{table}[h]
\begin{tabular}{cllll}
	\toprule
	 & \multicolumn{2}{c}{\bfseries Offloading Cost}&\multicolumn{2}{c}{\bfseries Revenue}\\\cmidrule{2-5}
	$k_v$&BKDtoRCS&PREDtoRCS&BKDtoRCS&PREDtoRCS\\ \midrule

20 &	2.72 $\pm$ 10.58 & 0.29 $\pm$ 1.17 & 14.56 $\pm$ 10.69 & 15.96 $\pm$ 2.22 \\

25 & 2.65 $\pm$ 10.8 & 0.33 $\pm$ 1.32 & 18.82 $\pm$ 10.88 & 20.29 $\pm$ 2.51 \\

30 & 2.86 $\pm$ 11.83 & 0.29 $\pm$ 1.2 & 22.88 $\pm$ 12.05 & 24.49 $\pm$ 2.86 \\

35 & 2.6 $\pm$ 12.28 & 0.26 $\pm$ 1.04 & 27.39 $\pm$ 12.61 & 28.67 $\pm$ 3.12 \\

40 & 2.43 $\pm$ 9.94 & 0.28 $\pm$ 1.08 & 31.91 $\pm$ 10.48 & 33.05 $\pm$ 3.45 \\

45 & 2.89 $\pm$ 14.73 & 0.28 $\pm$ 1.08 & 35.59 $\pm$ 15.19 & 37.29 $\pm$ 3.8 \\

50 & 2.99 $\pm$ 13.57 & 0.31 $\pm$ 1.22 & 39.83 $\pm$ 14.15 & 41.53 $\pm$ 4.09 \\

55 & 2.87 $\pm$ 15.09 & 0.29 $\pm$ 1.13 & 44.09 $\pm$ 15.83 & 45.7 $\pm$ 4.46 \\

60 & 2.84 $\pm$ 13.25 & 0.32 $\pm$ 1.23 & 48.43 $\pm$ 14 & 49.94 $\pm$ 4.8 \\

65 & 2.72 $\pm$ 14.28 & 0.31 $\pm$ 1.24 & 52.81 $\pm$ 15.09 & 54.27 $\pm$ 5.16 \\

70 & 3.04 $\pm$ 15.86 & 0.27 $\pm$ 1.1 & 56.77 $\pm$ 16.68 & 58.57 $\pm$ 5.41 
\\ \bottomrule
  \end{tabular}
  \caption{Average of offloading cost (the lower the better) and final revenue (the higher the better) for each test case, varying the plane capacity $k_v$ from 20m$^3$ to 70m$^3$. The model is designed is to reduce offloading costs and
the experiments show the results of the design.}
  \label{tbl:kv_results}
\end{table}
\fi

The left-hand side graph of Figure~\ref{fig:real_rev} clearly shows that for various capacity constraints ($k_v$) the offloading cost is lower almost by a factor of ten and with a much lower standard deviation. This suggests that using predictions instead of booked volume ($bkvol$) not only reduces the offloading cost but adds substantial amount of certainty into the whole air cargo booking process. 
In the right-hand side graph of Figure~\ref{fig:real_rev}, we show the final revenue, i.e.,~after subtracting the offloading costs, for various flight capacity constraints.
We observe that the revenue increases when using a predicted booked volume, albeit slightly, indicating that the decision function selected better-value shipments during the booking time horizon. Still the standard deviation of the revenue is lower when using the predictions. 
Note that, the way the decision making process is designed, excess overbooking incurs negative penalty (i.e.,~offloading), while underbooking results to zero penalty (i.e.,~$VF(x,0)$ is zero when the total volume is less than the flight capacity). It is thus more beneficial to reduce the risk of offloading by using a predictive model that overpredicts leading to less shipments getting accepted. This is a design choice that is driven by business objectives.

\subsection{In-depth analysis}
We now evaluate the main two components of our system independently: the predictive model and the decision making process.

\noindent \textbf{Predictive model:}
We first evaluate our predictor module.
For our evaluation, we use 3-fold cross-validation on the full real dataset of two years cargo bookings. We make a prediction on each single booking and we evaluate the aggregated flight leg predicted volume vs. the flight-leg received volume. For this reason, we implemented cross-validation so that all the bookings from the same flight leg are kept in the same split. Based on grid-search results, we set the XGBoost regressor with 0.9 subsample ratio of columns for each split, 300 estimators, a maximum tree depth of 20 and a learning rate of 0.05.  All other parameters are set as default.

\begin{figure}[h]
\includegraphics[width=1\columnwidth]{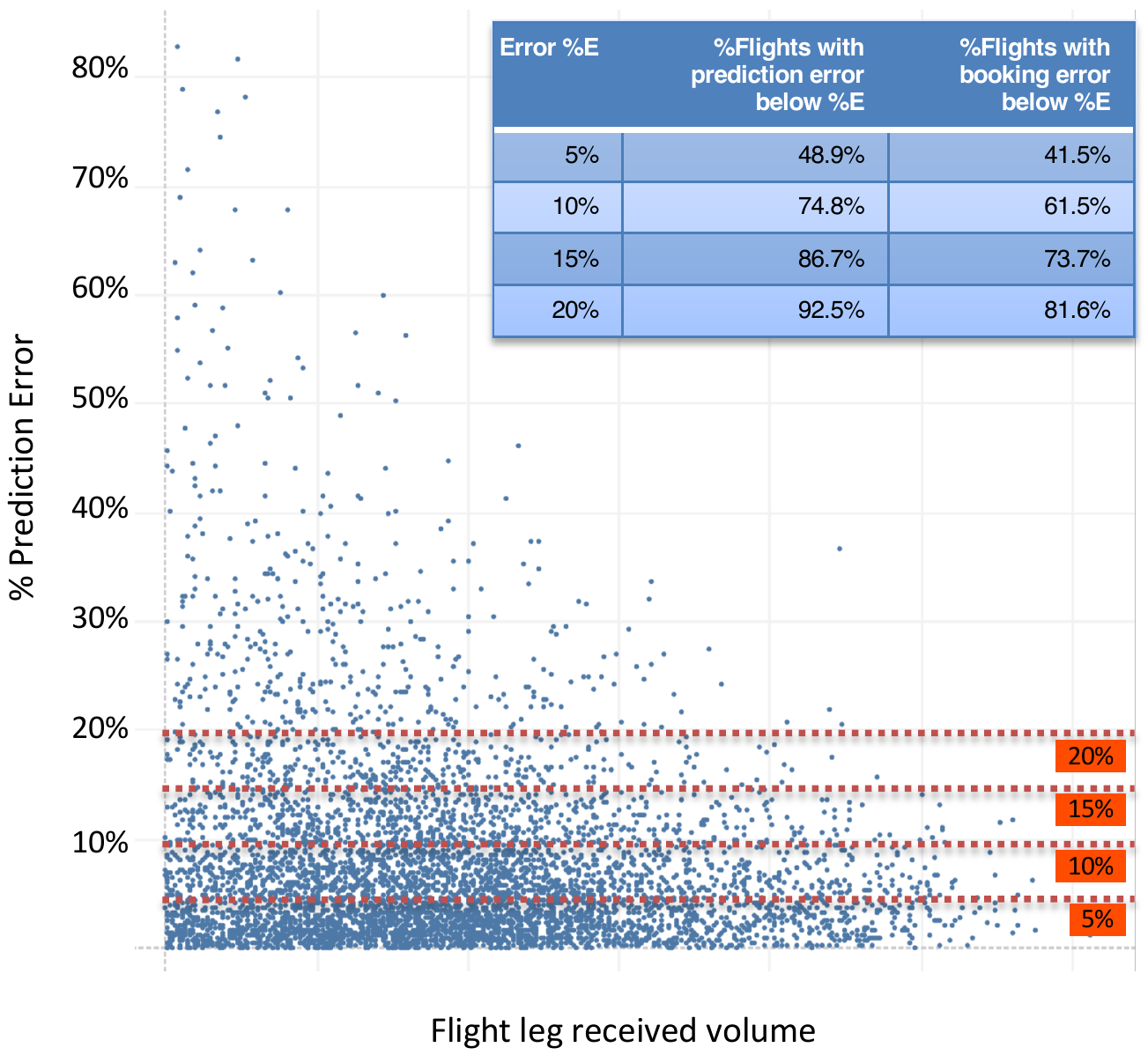}
\caption{Prediction error for flight legs in test splits. Each data point is a flight leg. Below each dotted line are all the flights with an error below the respective percentage.  }
\label{prediction_legs}
\end{figure}

We use the mean relative absolute error on our predictive model: $e=\frac{1}{N}\sum_1^N\frac{|rcsvol_i-f_\theta(bkvol_i)|}{rcsvol_i}$. The average error on the entire historical data is 7.8\%. Figure~\ref{prediction_legs} shows that the prediction error is under 5\% in almost half of the flights, while it is under 10\% for the 74.8\% of the flights. Similar results are shown in the table at the top right of the figure. We observe that by using the predictive model instead of the actual booked volume values we have a greater number of flights that have a small error. It is also noteworthy that our prediction error is lower for higher capacity flights, where it has the biggest impact.

%After our internal evaluation, the predictor module is now  deployed 
%in production and is validated over a period of two months. 

\begin{table}[h!]
\caption{Average error decrease for each product type, at booking level, in production environment.}
\label{tbl:prod_pred}
\begin{tabular}{crr}
	\toprule
	Product type & \% of booking& \begin{tabular}{@{}c@{}}\% error decrease  \\ from bkvol\end{tabular}\\\midrule
  Type1 & 73.3\% & -33.4\% \\
  Type2 & 10.1\% & -38.3\% \\
  Type3 & 3.7\% & -61.9\% \\
  Type4 & 3.5\% & -66.9\% \\
  Type5 & 3.1\% & -34.1\% \\
  Type6 & 2.7\% & -7.2\% \\
  Type7 & 1.4\% & -43.9\% \\
  Type8 & 0.8\% & -96.6\% \\
  Type9 & 0.7\% & +25.8\% \\
  Type10 & 0.1\% & -30.7\% 
  
\\\bottomrule
  \end{tabular}
\end{table}

Table~\ref{tbl:prod_pred} shows the benefits of the prediction on the shipment level for the 10 most frequent product types. 
Specifically, it shows the decrease in percentage of the prediction error (predicted $rcsvol$ vs. actual $rcsvol$) from the booking error (original booked volume vs. actual $rcsvol$). The predicted volume has decreased the error considerably for 9 out of the 10 product types. The increase in error for product type9 is 
on a very rare product type and, thus, there are not enough data to train the model. However, as it is rare it does not influence the total flight-leg predicted volume.

\noindent \textbf{Decision Making:}
We now evaluate our decision making module in isolation from the prediction module. This means that in this experiment we do not use predictions but the decision
is made based on the mean value $\bar{v_{i}}$ and offloading is determined based on the random number generated by a lognormal distribution, which leads to rule $D1S: R(\bar{v}_{i}) + VF(x + \bar{v}_{i},t-1) >  VF(x,t-1)$.
We evaluate $D1S$ by comparing it with a first-come first-served (FCFS) policy. In the FCFS policy, every incoming booking is 
accepted until the capacity runs out. FCFS is a greedy strategy in the
sense that it will accept immediate revenue instead of waiting for 
a potential booking from which more revenue can be made.  In an ideal 
setting, if the booking value was equal to the received value, 
FCFS will not incur any offloading costs. However, since the two
values are rarely the same, the natural advantage of FCFS is not 
realized and the experiments bear that out. Note 

\iffalse
try to maximize the revenue by accepting based on product type, in a perfect scenario the FCFS guarantees that there will be no offloading. However, since the received volume is usually different from the booked volume, this advantage is also lost.
\fi

We compare the $D1S$ and the FCFS strategy on the simulated data set from~\cite{amaruchkul2007single}.
Using the mean volume $\mu_k$ for  each product $k$ of the 24 categories 
 and the related probability distributions, we recreate synthetic data for 10 thousand different simulations and run the two different policies. Mean value is used as booked volume, while the received volume is drawn from a lognormal distribution with mean equal to the booked volume and variance $(\theta\mu_k)^2$.

\begin{table}[hbt]
	\caption{The average over 10,000 flight simulations for expected revenue and final revenue after offloading for the two strategies with varying $\theta$. Advantage in revenue for $D1S$ is kept after the offloading cost are applied.}
	\label{tbl:synth_results}
\begin{tabular}{ccc|cc}
	\toprule
	 & \multicolumn{2}{c}{\bfseries Expected revenue}&\multicolumn{2}{c}{\bfseries Revenue after offloading}\\\midrule
	$\theta$ & $D1S$ & FCFS & $D1S$ & FCFS\\ \cmidrule{2-5}

0.8 & 2927.87 & 2548.95 & 2584.28 & 2541.32\\

1.0 & 2927.87 & 2548.95 & 2553.74 & 2517.23

\\ \bottomrule
  \end{tabular}
\end{table}

Results in Table \ref{tbl:synth_results} show how $D1S$ has a benefit not only for the expected revenue but also the final one, after the offloading cost is deducted.
 As the
variance increases, offloading increases and revenue decreases which suggests
that our predictive approach has potential in better handling the booking decision making process, as we showed in the previous section.

\begin{table}[h!]
	 \caption{Last 16 time steps in the booking horizon of one flight simulation, showing current cargo load and decision for the two policies. }
	\label{tbl:heuristic_example}
\begin{tabular}{crrrrrr}
	\toprule
	 & & & \multicolumn{2}{c}{\bfseries FCFS}&\multicolumn{2}{c}{\bfseries $D1S$}\\\cmidrule{4-7}
 t & bkvol & \begin{tabular}{@{}c@{}}Revenue \\ rate\end{tabular}
  & Load & Decision & Load & Decision
 
 \\ \midrule

45 & 55.0 & 0.69 & 2759.0 & accepted & 2759.0 & accepted \\
46 & 59.0 & 0.79 & 2814.0 & accepted & 2814.0 & accepted \\
47 & 52.0 & 0.72 & 2873.0 & accepted & 2873.0 & accepted \\
48 & 30.0 & 1.12 & 2925.0 & accepted & 2925.0 & accepted \\
49 & 59.0 & 0.54 & 2955.0 & rejected & 2955.0 & rejected \\
50 & 119.0 & 0.98 & 2955.0 & rejected & 2955.0 & accepted \\
51 & 30.0 & 1.12 & 2955.0 & rejected & 3074.0 & accepted \\
52 & 30.0 & 1.12 & 2955.0 & rejected & 3104.0 & accepted \\
53 & 30.0 & 1.12 & 2955.0 & rejected & 3134.0 & accepted \\
54 & 52.0 & 0.8 & 2955.0 & rejected & 3164.0 & rejected \\
55 & 27.0 & 0.8 & 2955.0 & rejected & 3164.0 & rejected \\
56 & 52.0 & 0.8 & 2955.0 & rejected & 3164.0 & rejected \\
57 & 125.0 & 0.69 & 2955.0 & rejected & 3164.0 & rejected \\
58 & 30.0 & 1.12 & 2955.0 & rejected & 3164.0 & accepted \\
59 & 30.0 & 1.12 & 2955.0 & rejected & 3194.0 & accepted \\
60 & 30.0 & 1.12 & 2955.0 & rejected & 3224.0 & accepted

\\ \bottomrule
  \end{tabular}

\end{table}

Table \ref{tbl:heuristic_example} provides the last 16 steps in the booking horizon for one of the flight simulation.
In particular, it shows how in practice the decision rule $D1S$ keeps overbooking if the revenue rate is advantageous, while rejecting the less profitable shipments once the capacity is reached. This leads to the accumulation of offloading cost with the ultimate goal of maximizing the revenue.

\section{Discussion  and Future Work}
\label{section:discussion}
In this paper we have described the components of AI-CARGO, an intelligent
and data-driven air-cargo revenue management system. AI-CARGO was developed
in conjunction with a large commercial airliner over a two year period.
We summarize some of our lessons learned from the project and suggest directions
for future work. 

The project started as a classical data science project with the stated objective of predicting the received weight  (rcswt)
and received volume (rcsvol) of air-cargo bookings. We later narrowed it
to predicting rcsvol as it turned out that most flights are volume constrained,
i.e., they run out of volume space before weight capacity. However, overtime, we realized
that prediction per-se cannot be carried out in isolation. We have to analyze the 
upstream sources of data and understand how the data was being generated. Closer analysis led us
to conclude that the use of disguised missing values (DMVs) was prolific in the air-cargo
ecosystem and a solution was required to detect DMVs and obviate their impact on
the prediction task. We also had to get a better understanding of how the outcome
of the prediction task will be consumed by end users for decision making. This led us to formulate
the prediction-driven revenue optimization problem. Our general conclusion is that in order
to make real and tangible impact, data science techniques have to be situated and combined
with an overall objective. For future work we plan to extend the AI-CARGO system so
that it can be used by shippers and freight-forwarders and not just the RM teams
within an airline.

\end{document}